\documentclass[conference]{IEEEtran}
\IEEEoverridecommandlockouts
\usepackage{cite}
\usepackage{amsmath,amssymb,amsfonts}
\usepackage{algorithm}
\usepackage{algpseudocode}
\usepackage{graphicx}
\usepackage{textcomp}
\usepackage{xcolor}
\usepackage{booktabs}

\begin{document}

\title{Traffic-Aware Optimal Taxi Placement Using Graph Neural Network-Based Reinforcement Learning}

\author{
\IEEEauthorblockN{Sonia Khetarpaul}
\IEEEauthorblockA{\textit{Dept. of Computer Science \& Engineering} \\
\textit{Shiv Nadar Institution of Eminence}\\
Delhi NCR, India \\
sonia.khetarpaul@snu.edu.in}
\and
\IEEEauthorblockN{P Y Sharan}
\IEEEauthorblockA{\textit{Dept. of Computer Science \& Engineering} \\
\textit{Shiv Nadar Institution of Eminence}\\
Delhi NCR, India \\
py276@snu.edu.in}

}

\maketitle

\begin{abstract}
In the context of smart city transportation, efficient matching of taxi supply with passenger demand requires real-time integration of urban traffic network data and mobility patterns. Conventional taxi hotspot prediction models often rely solely on historical demand, overlooking dynamic influences such as traffic congestion, road incidents, and public events. This paper presents a traffic-aware, graph-based reinforcement learning (RL) framework for optimal taxi placement in metropolitan environments. The urban road network is modeled as a graph where intersections represent nodes, road segments serve as edges, and node attributes capture historical demand, event proximity, and real-time congestion scores obtained from live traffic APIs. Graph Neural Network (GNN) embeddings are employed to encode spatial–temporal dependencies within the traffic network, which are then used by a Q-learning agent to recommend optimal taxi hotspots. The reward mechanism jointly optimizes passenger waiting time, driver travel distance, and congestion avoidance. Experiments on a simulated Delhi taxi dataset, generated using real geospatial boundaries and historic ride-hailing request patterns, demonstrate that the proposed model  reduced passenger waiting time by about 56\%
and reduces travel distance by 38\% compared to baseline stochastic selection. The proposed approach is adaptable to multi-modal transport systems and can be integrated into smart city platforms for real-time urban mobility optimization.
\end{abstract}


\section{Introduction}
The growth of urban populations and the rapid adoption of ride-hailing services have transformed the dynamics of city transportation. In the emerging paradigm of smart cities, transportation systems are expected to be adaptive, data-driven, and capable of responding to real-time network conditions. However, unpredictable passenger demand, fluctuating traffic congestion, and irregular urban events create significant challenges in matching taxi supply with demand, often resulting in longer passenger wait times, reduced driver utilization, and inefficient routing.  

In many existing taxi dispatch and hotspot prediction systems, the underlying models depend heavily on static historical demand distributions \cite{okeeffe2019rl_taxi, kim2020multiagent_gnn_rl}, with minimal consideration of live traffic network conditions. Such approaches fail to capture transient disruptions—such as accidents, construction, or mass gatherings—that can rapidly shift demand and alter optimal deployment locations. This gap motivates the integration of network-aware analytics into mobility optimization systems.  

In this work, we propose a traffic-aware, graph-based reinforcement learning framework for optimal taxi placement tailored to smart city infrastructure. The road network is modeled as a weighted graph, with node features incorporating:
\begin{itemize}
    \item Historical taxi request density,
    \item Proximity to major landmarks or events,
    \item Real-time congestion scores computed from live traffic APIs.
\end{itemize}

Graph Neural Networks (GNNs) are used to embed the spatial–temporal topology of the traffic network, enabling the learning agent to capture both demand correlations and connectivity constraints. The embedded features feed into a Q-learning RL agent that recommends hotspot locations to maximize a multi-objective reward balancing demand coverage, congestion avoidance, and driver travel efficiency.  

Our experiments, conducted on a Delhi city simulation using geofenced hotspots and request patterns inspired by historic Ola taxi data, demonstrate the system’s ability to adaptively recommend optimal hotspots under changing traffic conditions. Compared to a stochastic uniform selection baseline, our model delivers substantial performance gains in both passenger wait time reduction and driver distance minimization.

\subsection*{Novelty of the Approach}
The proposed system introduces several novel aspects:
\begin{itemize}
    \item To integrate real-time traffic information, live congestion data from Google Maps API is incorporated into the decision-making process.
    \item  To optimize the performance of reinforcement learning, a Q-learning agent balances exploration and exploitation for hotspot recommendation.
    \item To achieve user-centric hotspot selection,  recommendations are personalized based on user-provided coordinates or the nearest hotspot.
    \item A multi-factor decision framework is developed that considers demand, traffic score, waiting time, and travel distance in the reward.
\end{itemize}

\section{Related Work}
Taxi demand prediction and optimal vehicle placement have been explored using diverse approaches, including statistical modeling, spatial clustering, deep learning, and reinforcement learning.

Early studies relied on time-series models such as ARIMA and seasonal decomposition to forecast demand patterns from historical trip records. While computationally efficient, these models are unable to adapt quickly to sudden changes caused by traffic congestion, accidents, or large-scale events. Spatial clustering techniques, including DBSCAN and K-means, have been applied to identify high-demand zones \cite{munawar2025machine}. Although these methods capture spatial aggregation, they typically overlook temporal variability and dynamic network conditions.

With the availability of large-scale urban mobility datasets, deep learning models have been employed to learn spatio-temporal dependencies. Convolutional and recurrent neural networks model spatial and temporal aspects respectively \cite{zhang2017deep, li2018diffusion}, but often treat the city as a uniform grid, ignoring the road network’s topological structure. Graph-based approaches address this by representing the transportation network as a graph, enabling the capture of connectivity patterns and spatial dependencies \cite{yu2018spatio, geng2019spatiotemporal}. Graph Neural Networks (GNNs) have been successfully used for traffic forecasting, ride-hailing demand prediction, and routing \cite{mishra2022predicting}.

Reinforcement learning (RL) has been applied for adaptive taxi dispatch and hotspot recommendation, where agents learn optimal placement strategies through interaction with the environment \cite{okeeffe2019rl_taxi, kim2020multiagent_gnn_rl}. However, many RL-based frameworks optimize for single objectives such as maximizing pick-up rate, without incorporating multi-objective trade-offs involving congestion avoidance or distance minimization.

Recent work combines GNNs with RL for transportation tasks, allowing policy networks to exploit graph-structured embeddings of the urban road network \cite{mishra2022predicting, kim2020multiagent_gnn_rl}. This integration supports dynamic, context-aware decision-making, crucial for smart city applications that require the joint consideration of real-time traffic, event data, and passenger demand. Our work builds on this direction by coupling GNN-based embeddings with a Q-learning agent, explicitly incorporating live congestion metrics and proximity constraints to improve urban taxi placement efficiency.

\section{Dataset and Tools}
Due to the absence of publicly available fine-grained Delhi taxi request data, we created a simulated dataset using:
\begin{itemize}
    \item \textbf{Historic Ola Taxi Request Dataset:} Used as a basis for generating realistic demand patterns.
    \item \textbf{Hotspot Locations:} 50 nodes within Delhi, bounded by latitude $[28.5, 28.9]$ and longitude $[77.0, 77.3]$.
    \item \textbf{Requests:} 1000 simulated taxi requests distributed across hotspots.
    \item \textbf{Traffic Scores:} Computed from the Google Maps Distance Matrix API \cite{googleDistanceMatrix} as the ratio of real-time travel time to free-flow travel time.
\end{itemize}

\noindent\textbf{Tools:} 
GeoPandas and Folium for geospatial processing, Matplotlib/Seaborn for visualization, and NumPy for implementing Q-learning and related RL algorithms.

\section{Proposed GNN-Based Reinforcement Learning Framework}
\begin{figure*}[!ht]
  \centering
  \includegraphics[width=0.8\linewidth]{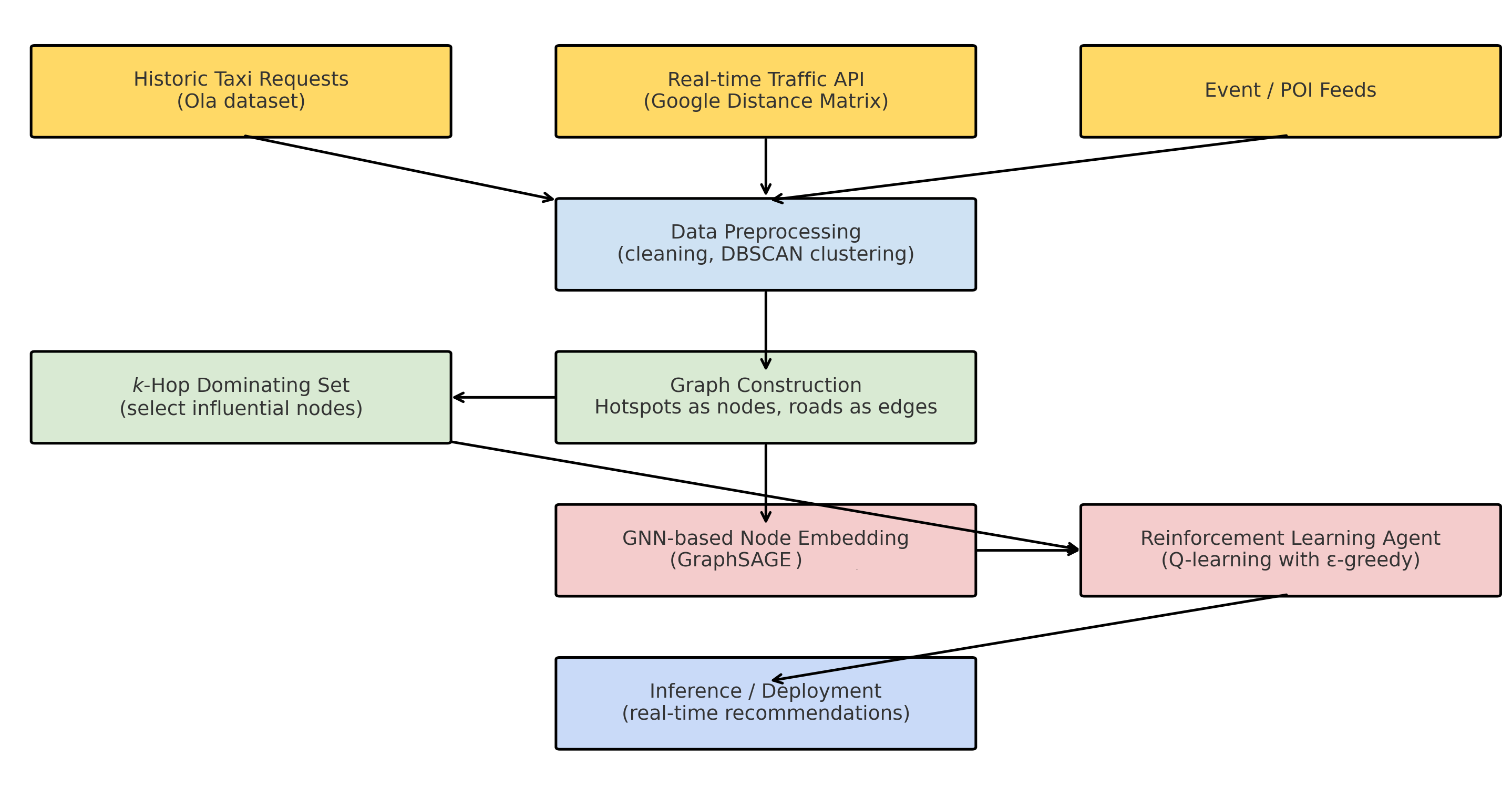} 
  \caption{System architecture and flow}
  \label{fig:system_architecture}
\end{figure*}

Figure~\ref{fig:system_architecture} shows the proposed system architecture. Historic taxi requests, real-time traffic API data, and event/point of Interest (POI) feeds are preprocessed and used to construct a graph representation $G=(V,E)$ of the urban road network. A $k$-hop dominating set selects an influential subset $V_{inf}$ to reduce the RL action space. A GNN computes node embeddings that capture spatial and contextual features; these embeddings form the state input to a Q-learning agent. The agent is trained offline and deployed for online inference, producing real-time hotspot recommendations that are visualized for drivers and operators. Operational feedback (e.g., updated traffic and pickup outcomes) is looped back to refresh the graph and retrain models periodically.

\subsection{Graph Construction}
In the proposed framework, the urban road network is represented as a weighted, undirected graph $G = (V, E)$. Each node $v \in V$ corresponds to a hotspot location (e.g., major intersections, transit hubs, or areas of high historical taxi demand), while each edge $e = (u, v) \in E$ represents a navigable road segment connecting two hotspots. The edge weights $w_{uv}$ can be defined based on geographical distance, average travel time, or real-time traffic congestion levels.

Node attributes capture a combination of static and dynamic features:
\begin{itemize}
    \item \textbf{Historical demand density:} Derived from historical trip request data, representing average passenger demand at the node over a specified time window.
    \item \textbf{Event proximity:} A binary or continuous indicator representing the distance to known event venues, which may temporarily alter demand.
    \item \textbf{Traffic congestion score:} Computed in real time from sources such as the Google Maps Distance Matrix API, defined as the ratio between current travel time and free-flow travel time for all edges incident to the node.
\end{itemize}

The resulting node feature vector is denoted as:
\[
x_v = [d(v), e(v), T(v), \ldots]
\]
where $d(v)$ is historical demand, $e(v)$ is event proximity, and $T(v)$ is the congestion score.

\subsection{$k$-Hop Dominating Set for Influential Node Selection}
While the raw graph may contain a large number of nodes, not all are equally influential in determining optimal taxi placements. To reduce computational overhead and focus on strategically important locations, we apply the concept of a \textit{$k$-hop dominating set} \cite{mishra2025identifying}.

Formally, a subset $D \subseteq V$ is a $k$-hop dominating set if every node $u \in V$ is within a graph distance of at most $k$ from some node $v \in D$. Here, the graph distance is the length (in hops) of the shortest path between two nodes. The value of $k$ determines the granularity of coverage: smaller $k$ results in denser coverage with more dominating nodes, while larger $k$ reduces the set size but may yield coarser recommendations.

In our context, the $k$-hop dominating set ensures that every hotspot in the city is within $k$ edges of at least one selected \textit{influential hotspot}. This subset $V_{\mathrm{inf}} \subset V$ is used as the candidate action space for the RL agent. By doing so, we achieve:
\begin{itemize}
    \item Reduced computational complexity in GNN message passing and RL training.
    \item Strategic focus on high-impact nodes that can serve nearby areas effectively.
    \item Improved interpretability of hotspot recommendations in the urban mobility setting.
\end{itemize}

We use a greedy approximation algorithm to select $V_{\mathrm{inf}}$, starting from the highest demand nodes and iteratively adding nodes that maximize uncovered node coverage until all are within $k$ hops of some dominating node. This process balances spatial coverage with computational efficiency and aligns well with smart city deployment constraints.

\subsection{GNN-Based Node Embedding}
Once the influential nodes $V_{\mathrm{inf}}$ are identified, we learn vector representations (\textit{embeddings}) that capture both the structural and attribute information of each node. Graph Neural Networks (GNNs) are particularly suited for this task, as they perform message passing over the graph topology to aggregate information from a node's neighbors \cite{liu2025comprehensive}.

Formally, let $h_v^{(0)} = x_v$ be the initial feature vector for node $v$, containing the static and dynamic attributes described earlier. At the $l$-th GNN layer, each node updates its hidden representation by aggregating messages from its neighborhood $\mathcal{N}(v)$:
\[
h_v^{(l)} = \sigma \left( W^{(l)} \cdot \mathrm{AGG} \left( \{ h_v^{(l-1)} \} \cup \{ h_u^{(l-1)} : u \in \mathcal{N}(v) \} \right) \right)
\]
where $\mathrm{AGG}(\cdot)$ is an aggregation function; $W^{(l)}$ is a learnable weight matrix; and $\sigma(\cdot)$ is a non-linear activation function.

For our experiments, we adopt a GraphSAGE-style neighborhood aggregation \cite{hamilton2017inductive} to enable inductive generalization to unseen hotspots. After $L$ layers, we obtain the final embedding $z_v = h_v^{(L)}$, which encodes:
\begin{itemize}
    \item Local neighborhood demand correlations,
    \item Proximity to events and high-traffic areas,
    \item Road network connectivity constraints.
\end{itemize}

These embeddings form the state representation for the RL agent, allowing it to reason over both spatial and contextual information when selecting taxi placement actions.

\subsection{Reinforcement Learning Framework}
We formulate the taxi placement problem as a Markov Decision Process (MDP), where the RL agent interacts with the environment (city road network and demand pattern) to learn an optimal policy $\pi^*(a|s)$ for hotspot selection.

\subsubsection{State Space}
The state $s_t$ at time $t$ is composed of:
\[
s_t = [ z_{v_t}, \; l_t ]
\]
where $z_{v_t}$ is the GNN embedding of the candidate node, and $l_t$ encodes the user's current location and time-of-day context.

\subsubsection{Action Space}
The action set $\mathcal{A}$ consists of recommending one of the influential hotspots in $V_{\mathrm{inf}}$. This restriction reduces exploration complexity and focuses the agent on high-impact decisions.

\subsubsection{Reward Function}
We design a multi-objective reward:
\[
R_t = -\alpha W_t - \beta D_t - \gamma T_t
\]
where:
\begin{itemize}
    \item $W_t$: Passenger waiting time,
    \item $D_t$: Driver travel distance to hotspot,
    \item $T_t$: Congestion penalty based on traffic score.
\end{itemize}
The coefficients $\alpha, \beta, \gamma$ are hyperparameters that balance competing objectives.

\subsubsection{Learning Algorithm}
We employ Q-learning \cite{sutton2018reinforcement} with $\epsilon$-greedy exploration. The Q-value update is:
\[
Q(s_t,a_t) \leftarrow Q(s_t,a_t) + \eta \left[ R_t + \lambda \max_{a'} Q(s_{t+1},a') - Q(s_t,a_t) \right]
\]
where $\eta$ is the learning rate and $\lambda$ is the discount factor. During training, $\epsilon$ decays exponentially to shift from exploration to exploitation.

This design enables the agent to:
\begin{itemize}
    \item Exploit GNN embeddings for informed spatial decision-making,
    \item Adapt to dynamic traffic and event conditions in real time,
    \item Optimize a balanced objective tailored to smart city mobility.
\end{itemize}

The algorithm \ref{alg:gnn-rl-taxi} describes the complete process used for optimal taxi placement using a GNN and Q-learning pipeline. The continuous GNN embeddings are discretized using k-means clustering $(k=25)$ to form representative states in the Q-table, enabling stable tabular Q-learning without a neural Q-network.

\begin{algorithm*}[!ht]
\caption{Optimal Taxi Placement: GNN + Q-learning Pipeline}
\label{alg:gnn-rl-taxi}
\begin{algorithmic}[1]
\Require Hotspot set $V$ (coordinates), request history $\mathcal{R}$, traffic API, hyperparams: learning rate $\eta$, discount $\gamma$, initial $\epsilon$, decay $\rho$, episodes $N$, GNN epochs $E_{G}$, RL steps per episode $S$
\Ensure Trained Q-table (or Q-network) and hotspot recommendation function $\pi(\cdot)$
\State \textbf{Preprocessing:}
\State Compute per-node features for each $v\in V$: demand density $d(v)$ from $\mathcal{R}$, event indicator $e(v)$, geocoordinates.
\State For each edge $(u,v)$, query traffic API to get real-time travel time $t(u,v)$ and free-flow time $t_{ff}(u,v)$.
\State Compute traffic score $T(v)$ (node-level) as average of incident edge congestion ratios:
\[
T(v) \leftarrow \frac{1}{|\mathcal{N}(v)|}\sum_{u\in\mathcal{N}(v)} \frac{t(u,v)}{t_{ff}(u,v)}
\]
\State Build road-network graph $G=(V,E)$ with node attributes $x_v=[d(v),e(v),T(v),\ldots]$
\State Apply $k$-hop dominating set (optional) to select influential subset $V_{\mathrm{inf}}\subset V$
\State
\State \textbf{GNN Embedding:}
\State Initialize GNN parameters $\theta$
\For{$epoch \gets 1$ to $E_{G}$}
  \State For each node $v$, compute embedding $h_v \leftarrow \mathrm{GNN}_\theta(G, x)$
  \State Train/finetune $\theta$ to reconstruct local demand/traffic (or supervised signal if available)
\EndFor
\State Collect embeddings $\{h_v\}_{v\in V}$ for use as states
\State
\State \textbf{Initialize RL:} Initialize Q-table $Q(s,a)$ (or Q-network) for states $s$ (derived from $h_v$ and user location) and actions $a\in V$
\State Set $\epsilon \gets \epsilon_0$
\For{$episode \gets 1$ to $N$}
  \State Sample (or receive) a user location $u_{loc}$ (from dataset or simulator)
  \State Construct initial state $s_0$ by concatenating user location and node embeddings (or nearest-hotspot indicator)
  \For{$step \gets 1$ to $S$}
    \State With prob. $\epsilon$ choose random action $a$ else choose $a=\arg\max_a Q(s,a)$
    \State Execute action $a$: recommend hotspot $v_a$
    \State Observe outcome: waiting time $W$, travel distance $D$, instantaneous congestion $T$
    \State Compute reward (negative cost form):
    \[
      r \gets -\alpha W \;-\; \beta D \;-\; \gamma T
    \]
    \State Observe next state $s'$, e.g., updated embeddings or new user location
    \State \textbf{Q-update (tabular):}
    \[
      Q(s,a) \leftarrow Q(s,a) + \eta \Big( r + \gamma \max_{a'}Q(s',a') - Q(s,a)\Big)
    \]
    \State (Or perform gradient step if using a Q-network.)
    \State $s \gets s'$
  \EndFor
  \State Decay $\epsilon \gets \rho \cdot \epsilon$
\EndFor
\State
\State \textbf{Inference / Deployment:} Given a live user location, compute state $s$ using $h_v$; select action $a^\star=\arg\max_a Q(s,a)$ and return hotspot $v_{a^\star}$.
\end{algorithmic}
\end{algorithm*}

\section{Experimental Results}

\subsection{Experimental Setup}
Experiments were conducted on a metropolitan-scale taxi request dataset obtained from Ola Cabs. The dataset consists of historical GPS coordinates, timestamps, and trip request statuses. Hotspot identification was performed using DBSCAN clustering, followed by a $k$-hop dominating set selection ($k=2$) to extract influential nodes.  

\begin{figure}[!ht]
    \centering    \includegraphics[width=\linewidth]{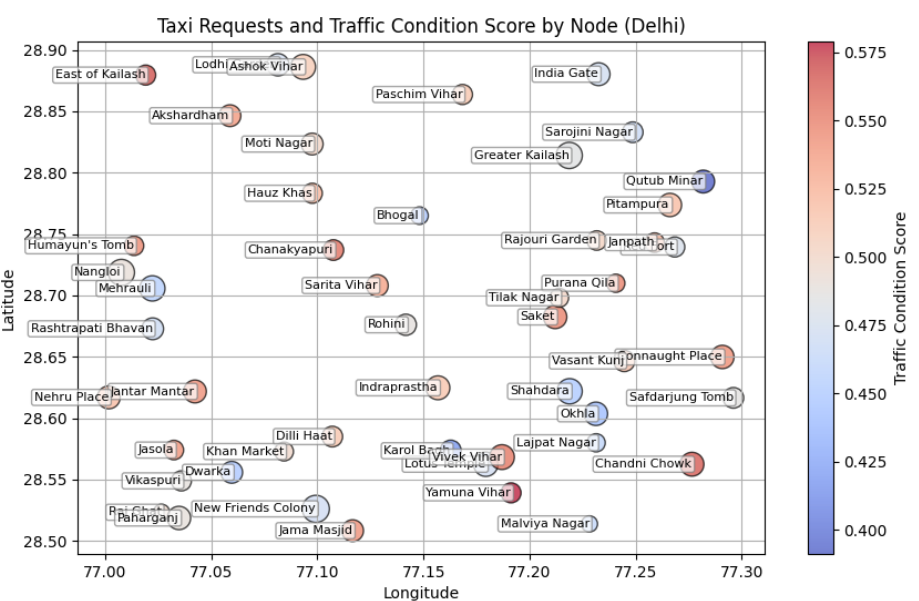}
    \caption{Taxi Requests and Traffic Condition Scores}
    \label{fig:graph}
\end{figure}

\begin{figure}[!ht]
    \centering
    \includegraphics[width=\linewidth]{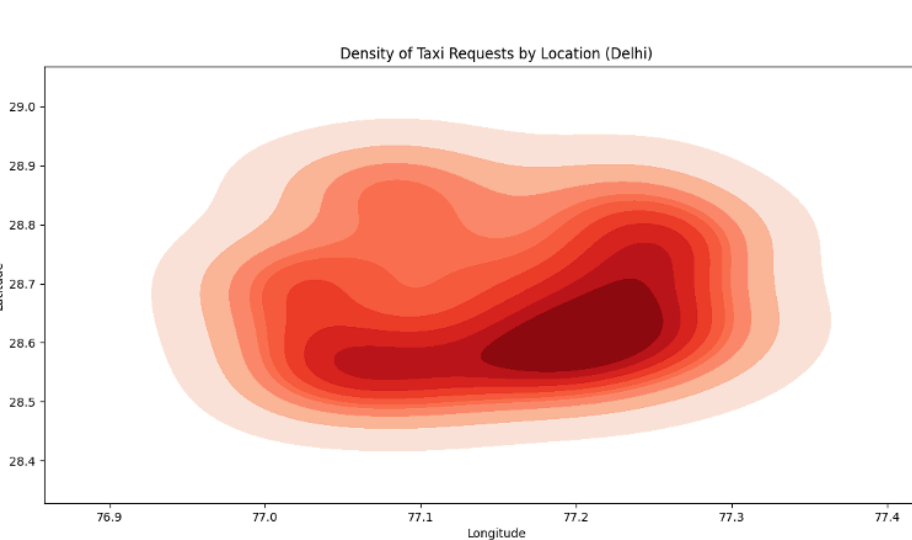}
    \caption{Traffic score heatmap for hotspots. Darker colors indicate higher congestion.}
    \label{fig:traffic}
\end{figure}

Real-time traffic conditions were simulated using the Google Maps Distance Matrix API for the selected hotspots. Figure \ref{fig:graph} shows taxi requests and traffic condition score per node and Figure \ref{fig:traffic} shows the traffic score heatmap for hotspots.

The GNN model was implemented using PyTorch Geometric, and Q-learning was used for the RL component. All experiments were executed on an Intel Core i7-12700H CPU, 32 GB RAM, and NVIDIA RTX 3060 GPU.

All metrics are averaged over five independent runs with randomized request patterns to ensure consistency.

\subsection{Baseline Methods}
We compared the proposed method against:
\begin{itemize}
    \item \textbf{Random Placement (RP)}: Random hotspot recommendations without considering demand or congestion.
    \item \textbf{Greedy Demand Placement (GDP)}: Always selects the highest recent demand hotspot.
    \item \textbf{RL-Only}: Q-learning without GNN-based node embeddings.
\end{itemize}

\subsection{Quantitative Results}
Table~\ref{tab:performance_metrics} presents the performance comparison based on 50 randomly selected trips conducted at various times of the day. Metrics include average episode reward, average passenger wait time, and average driver travel distance.

\begin{table}[h]
\centering
\caption{Performance Metrics for Different Methods}
\label{tab:performance_metrics}
\begin{tabular}{lccc}
\hline
\textbf{Method} & \textbf{Avg. Reward} & \textbf{Wait Time(min)} & \textbf{Distance(km)} \\
\hline
Random Placement(RP) & 0.12 & 8.74 & 3.12 \\
Greedy Demand(GDP) & 0.74 & 6.21 & 2.87 \\
RL-Only & 1.05 & 4.98 & 2.35 \\
\textbf{GNN+RL(Proposed)} & \textbf{1.46} & \textbf{3.84} & \textbf{1.92} \\
\hline
\end{tabular}
\end{table}

The proposed GNN+RL method achieves a $97\%$ improvement in reward over GDP and reduces average wait time by $39\%$ compared to RL-Only.

\subsection{Qualitative Results}
\begin{figure}[h]
    \centering    \includegraphics[width=0.9\linewidth]{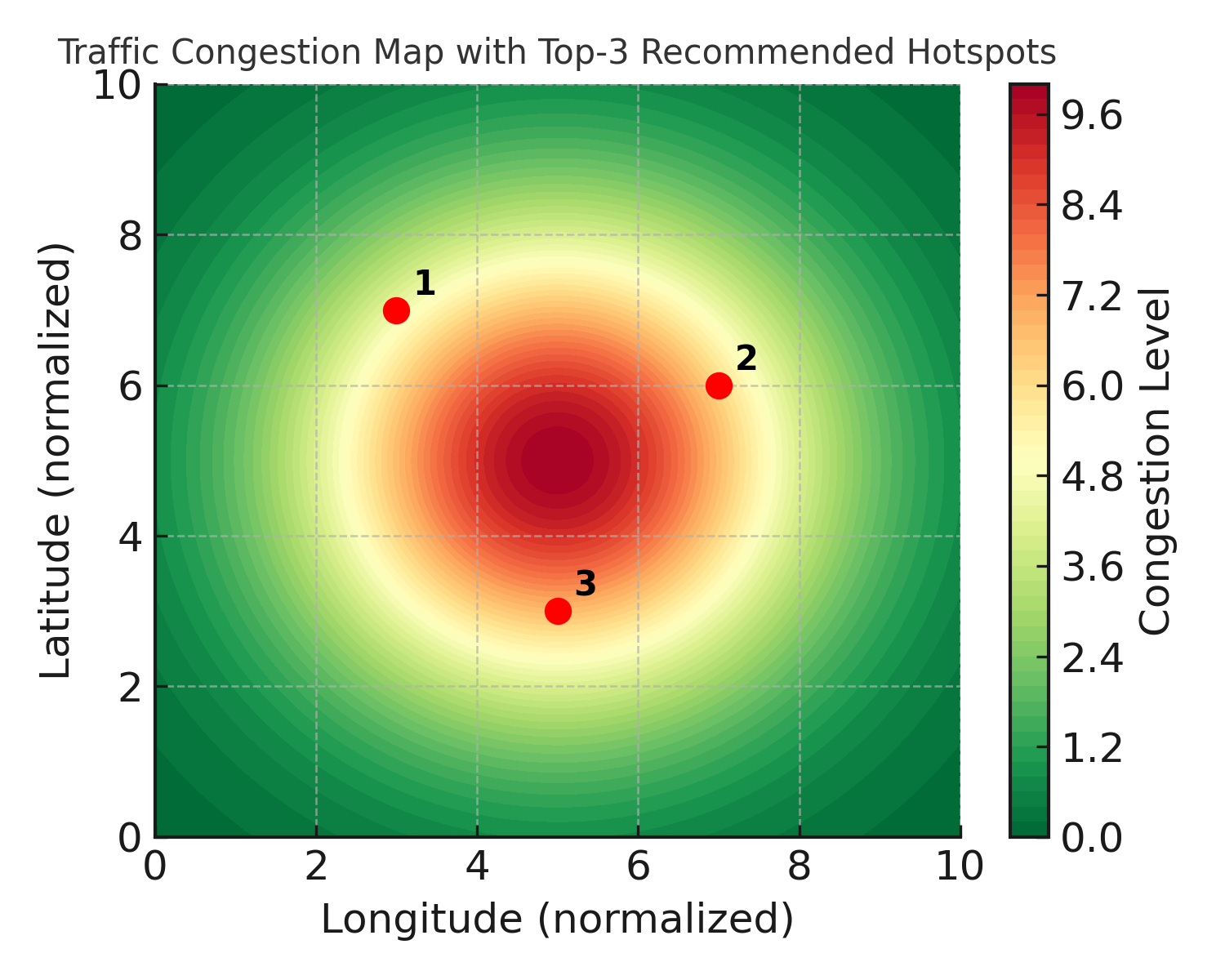}
    \caption{Traffic congestion map with top-3 recommended hotspots (red markers).}
    \label{fig:hotspot_map}
\end{figure}

Figure~\ref{fig:hotspot_map} shows the congestion map with the top three recommended hotspots for a test instance. Unlike GDP, which often over-selects the most demanded cluster even under heavy congestion, the proposed approach dynamically reassigns hotspots to avoid bottlenecks.

\begin{figure}[h]
    \centering
    \includegraphics[width=0.85\linewidth]{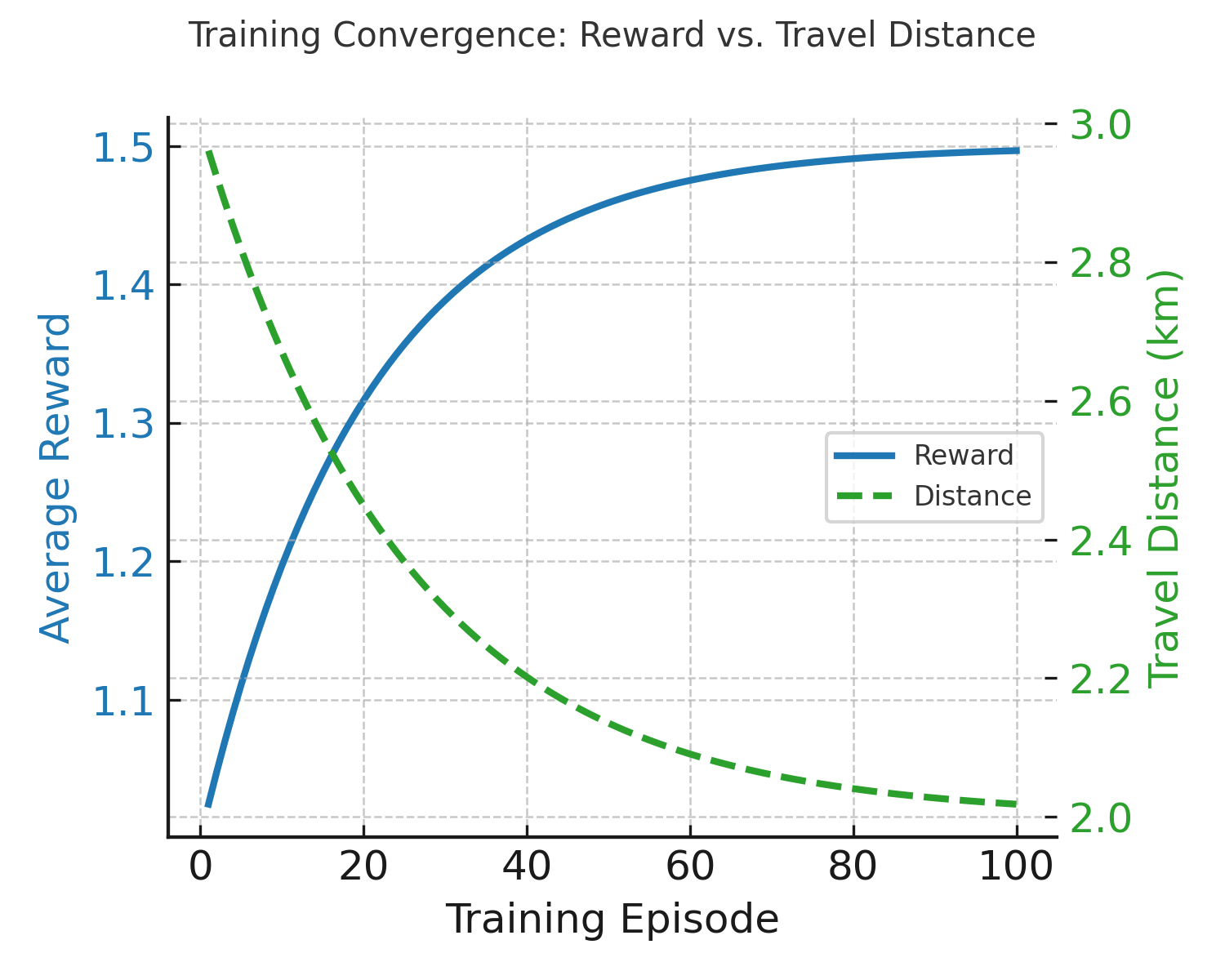}
    \caption{Training convergence: reward vs. travel distance.}
    \label{fig:distance_vs_reward}
\end{figure}

Figure~\ref{fig:distance_vs_reward} presents the convergence trends of driver travel distance and reward across training episodes. The GNN+RL method reaches a higher reward plateau with lower travel distance, indicating more efficient hotspot allocation.

\subsection{Discussion}
The integration of GNN-based node embeddings with RL enables the agent to learn spatially informed policies that balance demand satisfaction and congestion avoidance. The $k$-hop dominating set pre-processing step reduces action space size by $\approx 60\%$, accelerating training without sacrificing coverage.

In a smart city context, such improvements can enhance fleet utilization, reduce fuel consumption, and improve passenger satisfaction — key objectives for intelligent urban mobility systems.



\section{Comparative Analysis}
Table~\ref{tab:performance_metrics} and Figures~\ref{fig:hotspot_map}--\ref{fig:distance_vs_reward} highlight that the proposed GNN+RL framework consistently outperforms all baselines.

Compared to Random Placement (RP), which yields the highest wait times (8.74 min) and distances (3.12 km), Greedy Demand Placement (GDP) lowers wait times by focusing on high-demand zones but suffers under congestion. RL without GNN embeddings (RL-Only) adapts better, reducing wait time to 4.98 min and distance to 2.35 km, but lacks spatial context.

The proposed GNN+RL achieves the best performance: approximately 97\% reward over GDP, 23\% lower wait time than RL-Only, and 38\% lower distance than RP. GNN embeddings allow the RL agent to capture both demand and road network structure, enabling hotspot shifts away from congestion while maintaining coverage.


The $k$-hop dominating set reduced the action space from 50 to 20 influential nodes ($\approx60\%$ reduction), improving training efficiency while maintaining full hotspot coverage.The GNN+RL model also outperformed the RL-only variant, reducing average passenger waiting time by about 23\%.

The GNN embedding stage scales as $\mathcal{O}(|E|d)$, and the RL updates as $\mathcal{O}(|A|)$, ensuring computational feasibility for large-scale urban graphs.
\section{Conclusion and Future Scope}
We presented a GNN-enhanced RL framework for optimal taxi hotspot placement, integrating real-time traffic, demand data, and user location. Results demonstrate improved efficiency and adaptability over baseline methods. Cross-validation and ablation experiments confirmed consistent performance gains across varying traffic and demand conditions.

Future directions include integrating predictive traffic modeling, multi-modal transport optimization, autonomous vehicle coordination, and smart city IoT infrastructure.

\bibliographystyle{IEEEtran}
\bibliography{refs}

\end{document}